\documentclass[11pt,a4paper]{llncs}
\usepackage{graphicx}
\usepackage{float}

\usepackage{enumitem}
\usepackage{geometry}
\usepackage{hyperref}

\usepackage{cleveref}
\usepackage{paralist}
\usepackage{tabularx}
\usepackage{booktabs}
\usepackage{xcolor}

\usepackage[misc]{ifsym}
\title{\textbf{UNICORN: A Deep Learning Model for Integrating Multi-Stain Data in Histopathology}}

\author{Valentin Koch\textsuperscript{*,1,2}, Sabine Bauer\textsuperscript{*,3,5}, Valerio Luppberger\textsuperscript{4}, Michael Joner\textsuperscript{3,5}, Heribert Schunkert\textsuperscript{3,5},
Julia A. Schnabel\textsuperscript{1,2,6}, Moritz von Scheidt\textsuperscript{+,3,5}, Carsten Marr\textsuperscript{+,1}}
\institute{
\small\textsuperscript{1} Helmholtz Munich - German Research Center for Environmental Health, Munich, Germany\\
\small\textsuperscript{2} School of Computation and Information Technology, Technical University of Munich, Munich, Germany\\
\small\textsuperscript{3} Department of Cardiology, German Heart Centre Munich, TUM University Hospital, Munich, Germany\\
\small\textsuperscript{4} MLL Munich Leukemia Laboratory, Munich, Germany\\
\small\textsuperscript{5} DZHK (German Center for Cardiovascular Research), Partner Site Munich Heart Alliance, Munich, Germany\\
\small\textsuperscript{6} School of Biomedical Engineering and Imaging Sciences, King's College London, UK\\
 **,+ equal contribution
}

\begin{document}

\maketitle

\noindent \textbf{Background:} The integration of multi-stain histopathology images through deep learning poses a significant challenge in digital histopathology. Current multi-modal approaches struggle with data heterogeneity and missing data. This study aims to overcome these limitations by developing a novel transformer model for multi-stain integration that can handle missing data during training as well as inference. \\

\noindent \textbf{Methods:} We propose UNICORN (UNiversal modality Integration Network for CORonary classificatioN) a multi-modal transformer capable of processing multi-stain histopathology for atherosclerosis severity class prediction. The architecture comprises a two-stage, end-to-end trainable model with specialized modules utilizing transformer self-attention blocks. The initial stage employs domain-specific expert modules to extract features from each modality. In the subsequent stage, an aggregation expert module integrates these features by learning the interactions between the different data modalities. \\

\noindent \textbf{Results:} Evaluation was performed using a multi-class dataset of atherosclerotic lesions from the Munich Cardiovascular Studies Biobank (MISSION), using over 4,000 paired multi-stain whole slide images (WSIs) from 170 deceased individuals on 7 prespecified segments of the coronary tree, each stained according to four histopathological protocols. UNICORN achieved a classification accuracy of 0.67, outperforming other state-of-the-art models. The model effectively identifies relevant tissue phenotypes across stainings and implicitly models disease progression. \\

\noindent \textbf{Conclusion:} Our proposed multi-modal transformer model addresses key challenges in medical data analysis, including data heterogeneity and missing modalities. Explainability and the model's effectiveness in predicting atherosclerosis progression underscores its potential for broader applications in medical research. \\
\newpage
\section*{Introduction}
Atherosclerosis, a complex inflammatory disease of the arterial wall, is a leading cause of cardiovascular morbidity and mortality worldwide. Histopathological classification of atherosclerosis, as proposed by Stary et al. \cite{Stary1994-br,Stary1995-fz} and adapted by Virmani et al. \cite{Virmani2000-jr} and Otsuka et al. \cite{Otsuka2014-hm}, plays a crucial role in assessing disease severity, progression and potential therapeutic interventions. This classification scheme assesses features such as the thickness of the fibrous cap, the presence of a necrotic core, the degree of inflammation and the extent of calcification within arterial plaques \cite{onnis2024coronary,lee2021erratum,kassis2022fibrous}.
Whole slide imaging (WSI) allows detailed visualization of tissue samples at micrometer resolution, providing critical insight into various medical conditions. While the most commonly used staining method, haematoxylin and eosin (H\&E), provides a sufficient overview of the tissue and is used to diagnose many diseases, in some cases further staining protocols are required to fully understand underlying tissue characteristics\cite{Leong2009-mz}. For example, immunohistochemistry (IHC) staining is essential for cancer detection, providing critical prognostic, diagnostic, and therapeutic information that enables precise and personalized treatment strategies \cite{kim2016immunohistochemistry}. Specific staining methods, like von Kossa silver stain or Movat pentachrome stain, highlight particular tissue components, such as mineralisation or connective tissue composition and lipid distribution \cite{Jones2008-yn}.  Manual integration and interpretation of multi-stain data is challenging, as multiple high-resolution images that are each potentially gigabytes in size need to be examined in detail. 
In recent years, computational pathology, focusing on the analysis of digitized pathology slides, has seen remarkable advances, largely driven by modern machine learning techniques, especially deep learning. These advances cover several  areas including disease classification, tissue segmentation, and mutation prediction \cite{Wagner2023-qu,Van_der_Laak2021-ac,Chen2022-wn,Al-Kofahi2010-ug}. Given the large size of WSIs, a common strategy is to decompose them into manageable-sized image patches. Subsequently, a pre-trained feature extractor condenses each patch into a low-dimensional feature vector, which is then processed by a Multiple Instance Learning (MIL) network \cite{Ilse2018-do,shao2021transmil}. Recent developments in transformer architectures, in particular Vision Transformers (ViT), have shown promising results as the backbone of state-of-the-art feature extractors \cite{chen2024towards,campanella2023computational,vorontsov2024}, and when applied as WSI classification networks\cite{Wagner2023-qu}. Although there has been a lot of research on cancer prediction and using deep learning on WSIs, computational pathology is quite new in the field of atherosclerosis. Holmberg et al. used co-registered histopathology and OCT images to improve segmentation of calcification and lesions in OCT images using a UNet \cite{Holmberg2021-ak}. The open-source tool Vesseg has been developed to segment lesions using U-nets on H\&E stained brachiocephalic arteries\cite{Murray2021-ll}. 
In our work, we introduce UNICORN, a two-stage, end-to-end trainable transformer architecture capable of handling different modalities, in particular features obtained from WSIs with various staining protocols. To our knowledge, this is the first AI model tailored for multi-stain histopathological classification. We show how UNICORN implicitly captures disease progression in atherosclerosis, handles missing data effectively during both training and inference, demonstrating its robust applicability in real-world settings.

\begin{figure}[htbp]
    \centering
    \includegraphics[width=\textwidth]{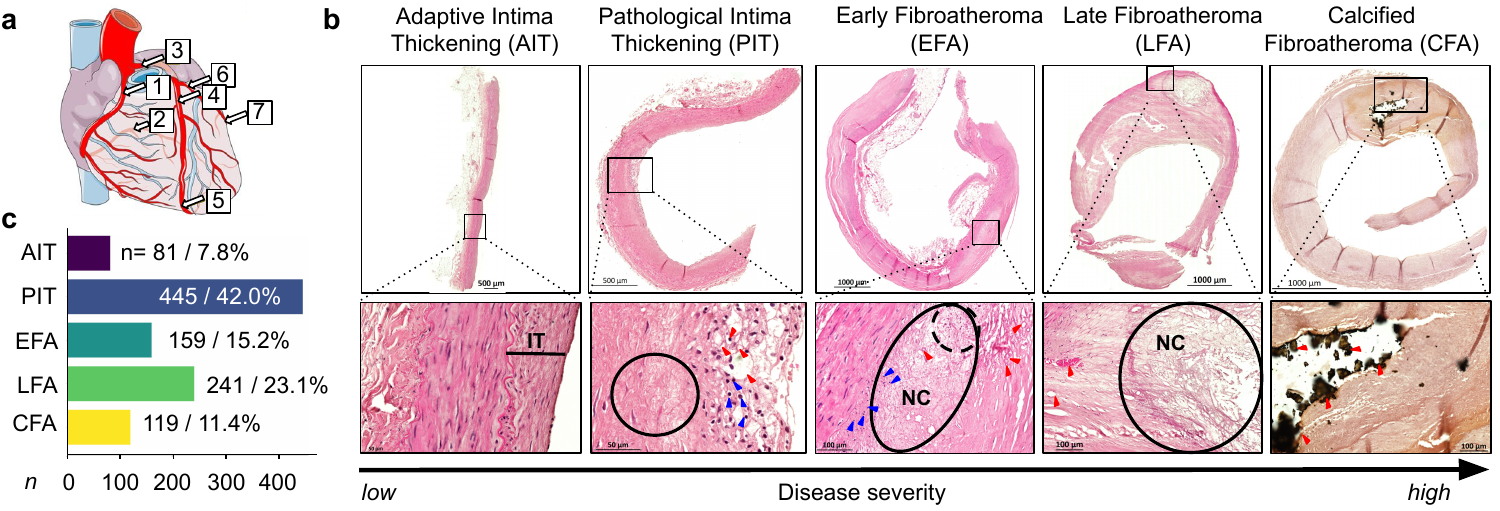}
    \includegraphics[width=\textwidth]{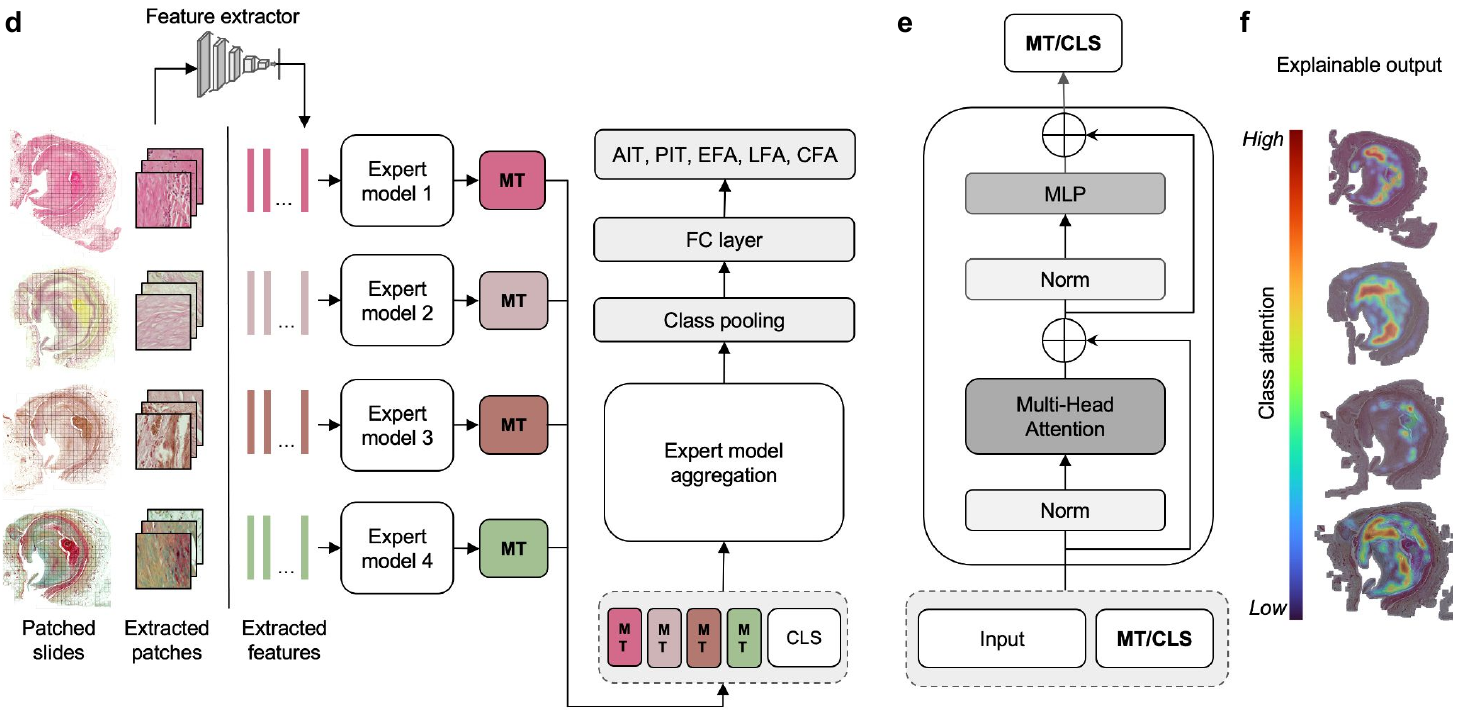}
    \caption{\textbf{MISSION biobank and UNICORN architecture.} \\
\textbf{a)} Coronary artery segments used in this study. 1+2: proximal and distal part of the right coronary artery, 3: main stem, 4+5: proximal and distal part of the left coronary artery, 6+7: proximal and distal part of the left circumflex coronary artery. \\
\textbf{b)} Histological classification of coronary arteries according to an adapted and simplified AHA classification based on Virmani et al. \cite{Virmani2000-jr}. H\&E stains are shown for AIT, PIT, EFA, LFA and von Kossa silver stain is shown for CFA, all exemplary with zoomed in regions of interest. Blue and red arrows show class specific characteristics.\\
\textbf{c)} Class distribution in the study cohort of MISSION with each n=7 segments of 170 individuals. \\
\textbf{d)} UNICORN architecture: features extracted from WSIs with the four different stainings Hematoxylin and Eosin (H\&E), Elastica van Gieson (EvG), von Kossa (vK) and Movat pentachrome (Movat) are forwarded to four expert models consisting of two transformer blocks (e) that are specialized in processing data from a certain staining. Similar to the class token (CLS), information from each domain is aggregated in a modality token (\textbf{MT}) which is used as input to the expert aggregation model, that combines information across stainings into the CLS token. The final classification score is derived from a fully connected (FC) layer that uses the CLS token as input. \\
\textbf{e)} Transformer blocks: transformer blocks used in expert models and expert aggregation models are shown in detail. \textbf{MLP}: multi-layer perceptron \\
\textbf{f)} Explainable output of UNICORN: using attention values, UNICORN can provide explainable output, highlighting regions of importance across stainings. \\
\textbf{IT}: Intima thickening, \textbf{NC}: necrotic core}
    \label{fig:fig1}
\end{figure}

\section*{Results}
\subsection*{UNICORN architecture}
UNICORN (UNiversal stain Integration network for CORonary classificatioN) is a network capable of integrating and processing heterogeneous data across different tissue stainings (figure \ref{fig:fig1}). The model is end-to-end trainable and exhibits resilience to data incompleteness during both training and inference. The network comprises specialized modules, each having the same architecture. These modules include a two-layer self-attention transformer block with four attention heads, as illustrated in figure \ref{fig:fig1}e. Initially, the input data is encoded: the entire slide image is tiled into 256x256 px sized patches. A pre-trained feature extractor is then used to generate embeddings for each patch (see Methods for details). Each staining is individually processed by specialized domain expert modules that learn unique domain-specific features (figure \ref{fig:fig1}d). Each of the expert modules propagates a modality token (MT), analogous to the class token (CLS) used in vision transformers, to the aggregation expert module. The aggregation expert module learns to aggregate the information from the different stainings in a CLS token, which is ultimately used as input to the fully-connected (FC) layer which outputs a classification score (figure \ref{fig:fig1}d). Since the aggregation expert module, like all expert modules, is a transformer that can handle a variable number of input tokens, it is capable of processing data with missing modalities both during training and inference. UNICORN is evaluated on a multi-stain, multi-class atherosclerosis dataset where it classifies five stages of coronary atherosclerosis. 
\subsection*{MISSION: Multi-stain atherosclerosis classification dataset}
The dataset consists of tissue sets from 170 deceased individuals. Each set contains 7 segments from different parts of the coronary tree (figure \ref{fig:fig1}a): proximal and distal part of the right coronary artery, main stem, proximal and distal part of the left coronary artery, proximal and distal part of the left circumflex coronary artery. Each segment is stained using four different staining methods: Hematoxylin and Eosin (H\&E), Elastica van Gieson (EvG), von Kossa (vK) and Movat pentachrome (Movat) stain. For lesion classification, a modified scheme according to the AHA and the classification method described by Virmani et al. \cite{Virmani2000-jr,Otsuka2014-hm} was employed: adaptive intima thickening (AIT), pathological intima thickening (PIT), early fibroatheroma (EFA), late fibroatheroma (LFA) and calcified fibroatheroma (CFA). These labels (see Methods and figure \ref{fig:fig1} for detailed description) have been assigned by two biomedical experts after reviewing all four stainings. Disease severity increases from AIT over PIT, EFA, and LFA to CFA. Class specific characteristics are highlighted in figure \ref{fig:fig1}b: AIT samples show intima thickening (IT) but intact cell structure and smooth muscle cell layer in a proteoglycan and collagen rich matrix. PIT is described as preatherosclerotic lesions with focal fat-laden macrophages (red triangles), inflammatory cells (blue triangles) and fatty streaks. The gray circle shows matrix remodeling resulting in a varying degree of smooth muscle cells. In the EFA sample, the black circle highlights the necrotic core (NC), the blue arrows show inflammatory cells and the red arrows point to the lipid pool. The black dotted circle shows cell debris. In the LFA case, the gray circle shows the NC with cholesterol crafts and fibrocalcific surrounding, the red arrows point to neovascularization. The red arrows in CFA show calcification in the necrotic core in the vK stained slide.

\begin{figure}[htbp]
    \centering
    \includegraphics[width=\textwidth]{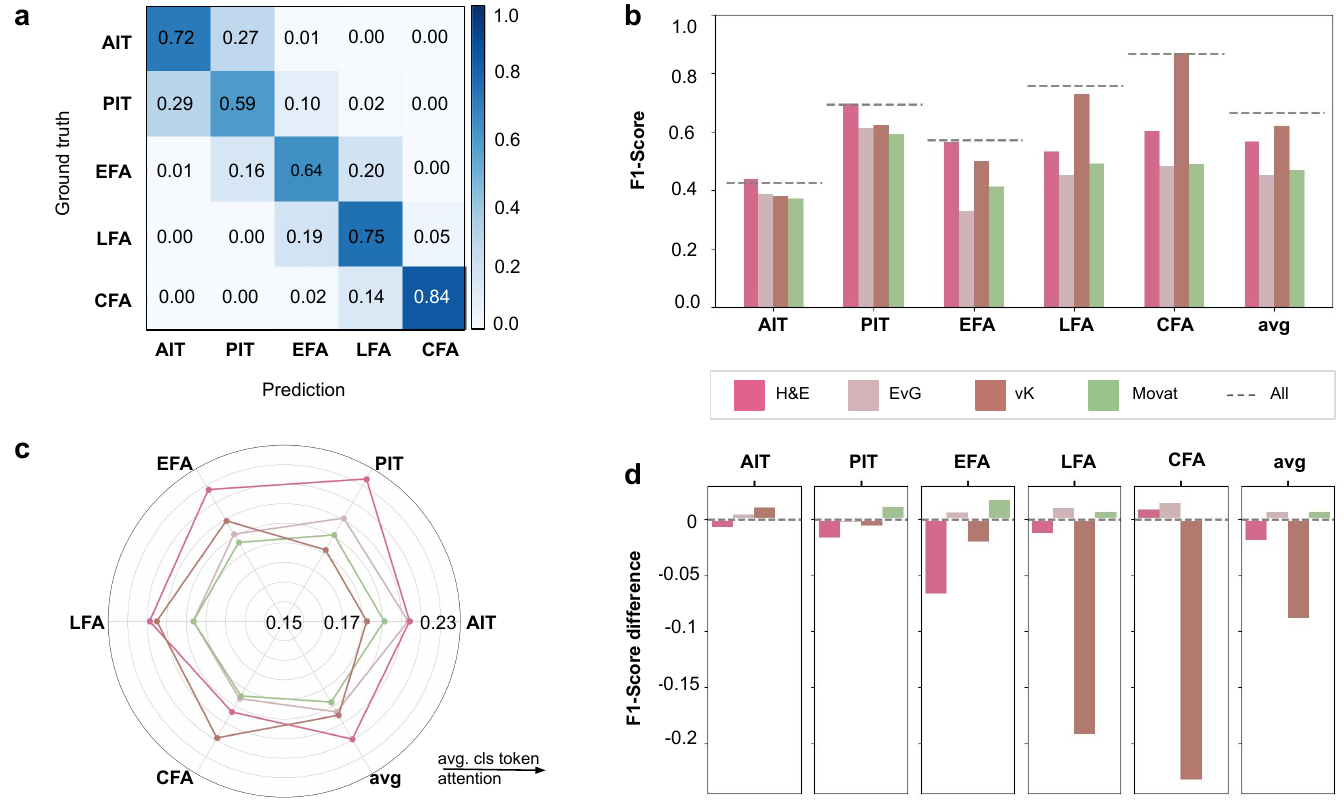}
    \caption{\textbf{UNICORN integrates information from four stainings.}\\
\textbf{a)} Confusion matrix shows UNICORN classifying multi-stain atherosclerosis WSIs into five classes with high accuracy. Values are color coded from white (0) to dark blue (1).\\
\textbf{b)}  Performance of UNICORN using just one staining as input. The highest F1-Score is achieved when using all four stainings (grey dotted line), indicating that UNICORN successfully aggregates information of the different stainings.\\
\textbf{c)}  The attention mechanism of the UNICORN aggregation expert module functions well, with higher attention values noted for stainings exerting a high influence on performance for a given class. The mean attention value of the CLS token to the four modality tokens corresponding to a given staining is shown.\\
\textbf{d)} F1-Score difference when utilizing three of four stainings vs. all stainings (gray dotted zero line), bar colors indicate which staining was excluded. Findings demonstrate robust correlation between two importance measurements (b, d) based on performance with the learnt attention scores (c) by UNICORN. 
}
    \label{fig:fig2}
\end{figure}

\subsection*{Evaluation}
UNICORN is evaluated in a 5-fold cross validation scheme, splitting the dataset into 60/20/20 train/validation/test set for each fold (see Methods for details). We show that the network outperforms simple approaches that do not take into account the multi-modality of the data (Table 1). Our model achieves an average F1-Score of 0.66 ± 0.04 (mean ± standard deviation) and an accuracy of 0.67 ± 0.05 compared to an F1-Score of 0.63 ± 0.05 and accuracy of 0.64 ± 0.05 for the second-best model.

\begin{table}[h!]
\centering
\begin{tabular}{l@{\hskip 0.5cm}c@{\hskip 0.5cm}c}
\textbf{Model} & \textbf{F1-Score} & \textbf{Accuracy} \\ \hline
AttentionMIL \cite{Ilse2018-do}  & 0.43 ± 0.03 & 0.55 ± 0.02 \\ 
Perceiver \cite{jaegle2021perceiver}     & 0.53 ± 0.08 & 0.56 ± 0.11 \\ 
Transformer \cite{Wagner2023-qu}   & 0.63 ± 0.05 & 0.64 ± 0.05 \\ \hline
\textbf{UNICORN}            & \textbf{0.66 ± 0.04} & \textbf{0.67 ± 0.05} \\ 
\end{tabular}
\caption{Performance comparison between UNICORN and three other multiple instance models on multi-stain coronary artery classification.}
\end{table}

Most misclassifications are observed between adjacent disease stages, which are inherently difficult to classify and distinguish even for highly trained experts (figure \ref{fig:fig2}a). Our model demonstrates strong performance with single-staining inputs, but achieves optimal results when utilizing all available staining methods (figure \ref{fig:fig2}b), demonstrating its capabilities of aggregating information. When using only one staining as input, the vK staining performs best in classifying CFA and LFA and second best on EFA. Using H\&E staining works best for the classes AIT, PIT and EFA. Movat and EvG perform roughly on par across classes (figure \ref{fig:fig2}b). 
The attention of the CLS token to the stain specific modality token in the expert aggregation model (figure \ref{fig:fig2}c) demonstrates that the model effectively focuses on the most relevant staining for its predictions. H\&E staining is predominantly given the highest attention, except for LFA and CFA, where vK staining is essential, aligning well with findings from figure \ref{fig:fig2}b. EvG is given slightly higher attention than Movat on AIT and PIT, again in concordance with \ref{fig:fig2}b, where EvG staining performs slightly better on these classes.
In figure \ref{fig:fig2}d, one staining is left out at inference to visualize the difference in accuracy compared to inference with all four stainings. This gives another measurement of how important a specific staining is to the model's performance. The results again align well with the other experiments: Excluding von Kossa silver stain reduces the classification performance in advanced phenotypes (LFA, CFA), while H\&E is the most important staining for the remaining classes. Removing EvG or Movat has little effect and sometimes even improves performance, potential explanations are discussed below.  

The results presented in figures \ref{fig:fig2}b-c are not only consistent with one another, but also illustrate the relevance of the different stainings for manual characterization and the underlying biological processes. For instance, the vK staining, which imparts a black chromatic hue to calcifications, is of particular importance for the identification and differentiation of the calcified fibroatheroma (CFA) and lipid-rich late fibrous atheromas (LFA). As also illustrated in figures \ref{fig:fig2}b and \ref{fig:fig2}c, H\&E staining remains a fundamental tool for evaluating cell densities and general tissue structure, which is the most crucial aspect for differentiating AIT and PIT. AIT and PIT represent different stages of atherogenesis, with variations in cell density, inflammatory infiltration, and extracellular matrix organization. 

\begin{figure}[htbp]
    \centering
    \includegraphics[width=\textwidth]{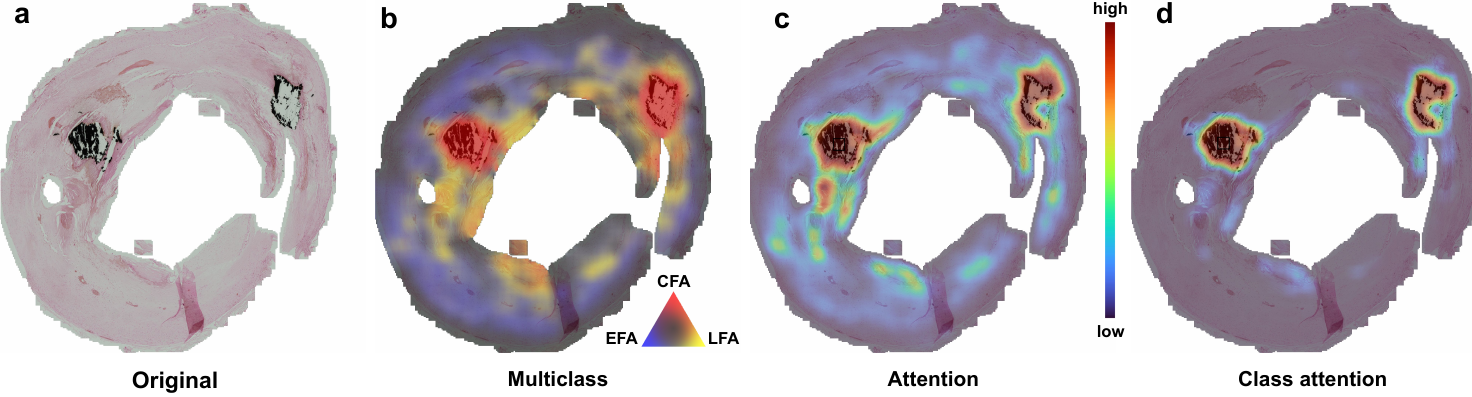}
    \caption{\textbf{High resolution heatmaps highlight relevant regions for classification decisions.}\\
\textbf{a)} Original von Kossa silver stain image showing black calcification regions.\\
\textbf{b)}  Coloring indicates the presence of structures associated with the most severe disease classes (EFA=blue, LFA=yellow, CFA=red).\\
\textbf{c)}  Attention scores (lower score = blue, high score/high importance= red) show high attention regions.\\
\textbf{d)} Multiplication of attention score (lower score = blue, high score/high importance = red) by the probability of the class that is predicted by UNICORN illustrates which region UNICORN considers important and also suspects the class it predicts to be in.
}
    \label{fig:fig3}
\end{figure}

\subsection*{UNICORN highlights explainable features}
Three different visualization methods are provided to improve explainability: the typical attention mechanism, resulting from an attention rollout \cite{abnar2020quantifying} over both stages of the transformer (figure \ref{fig:fig3}). Figure \ref{fig:fig3}b illustrates a tissue phenotype visualization, where tissue regions are highlighted based on the model's interpretation of the class most indicative of each area (see Methods). Due to practical reasons, in this case only the three most progressed disease tissue types are considered. Figure \ref{fig:fig3}c highlights areas of attention computed by attention rollout independently per staining. Figure \ref{fig:fig3}d combines the information of the first two methods into "class attention," which highlights the tissue phenotypes that the model identifies with high attention and associates with its predicted class. In conclusion, three meaningful visualization maps are given: a) “Where does the model attend to?” b) “Where does the model find phenotypes of the respective classes”, and c) “Where does it attend to the predicted class” which in our experiments highlights the areas of interest best and which are used in the following figures. These visualizations improve the interpretability of the model's decisions significantly, allowing researchers and clinicians to map the model’s predictions to specific tissue regions. This method not only helps confirm the accuracy of the model’s outputs with recognized pathological features but also increases confidence in the model by transparently highlighting its decision-making process.

\begin{figure}[htbp]
    \centering
    \includegraphics[width=\textwidth]{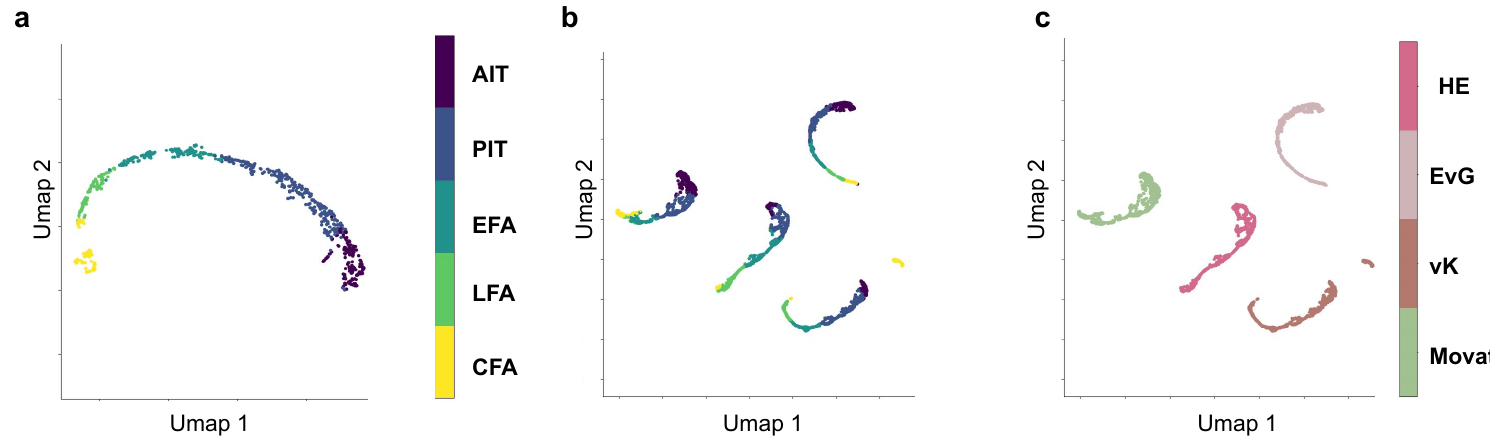}
    \caption{\textbf{UMAP reveals disease progression modeling capabilities.}\\
UMAP of features of the final layer illustrates that the model learns the natural disease progression from AIT to CFA (a) implicitly. This finding is consistent when using only one staining as an input (b), and the model is able to distinguish which staining it is processing (c).
}
    \label{fig:fig4}
\end{figure}

Moreover, figure \ref{fig:fig4}a illustrates that UNICORN is capable of accurately modeling the natural progression of atherosclerosis, as evidenced by a UMAP of an additional internal dataset comprising 768 segments from 114 individuals. The capacity of UNICORN to capture this progression is particularly noteworthy as it implies that the model is not only performing classification but is also learning a representation that aligns with the underlying biological process of the disease. This suggests that the model's decision-making is grounded in clinically relevant features, which could enhance its utility in both diagnostic and research settings.
Furthermore, the progression model exhibits consistency when considering individual stainings, as shown in figure \ref{fig:fig4}b. The model's robustness when limited to a single staining demonstrates its adaptability and resilience in scenarios where multimodal data might not be available. This means that the model can still provide meaningful insights even when only partial data is accessible, which is common in real-world clinical environments. The ability to generalize to a single staining suggests that the features learned by the model are rooted in the biological characteristics of the tissue, rather than being artifacts of the specific staining technique. As illustrated in figure \ref{fig:fig4}c, the model’s capacity to differentiate between various stainings is supported by empirical evidence. This ability to distinguish the particular staining processing is a crucial step towards effectively handling missing data, as it inherently recognizes what information is absent.

\begin{figure}[htbp]
    \centering
    \includegraphics[width=\textwidth]{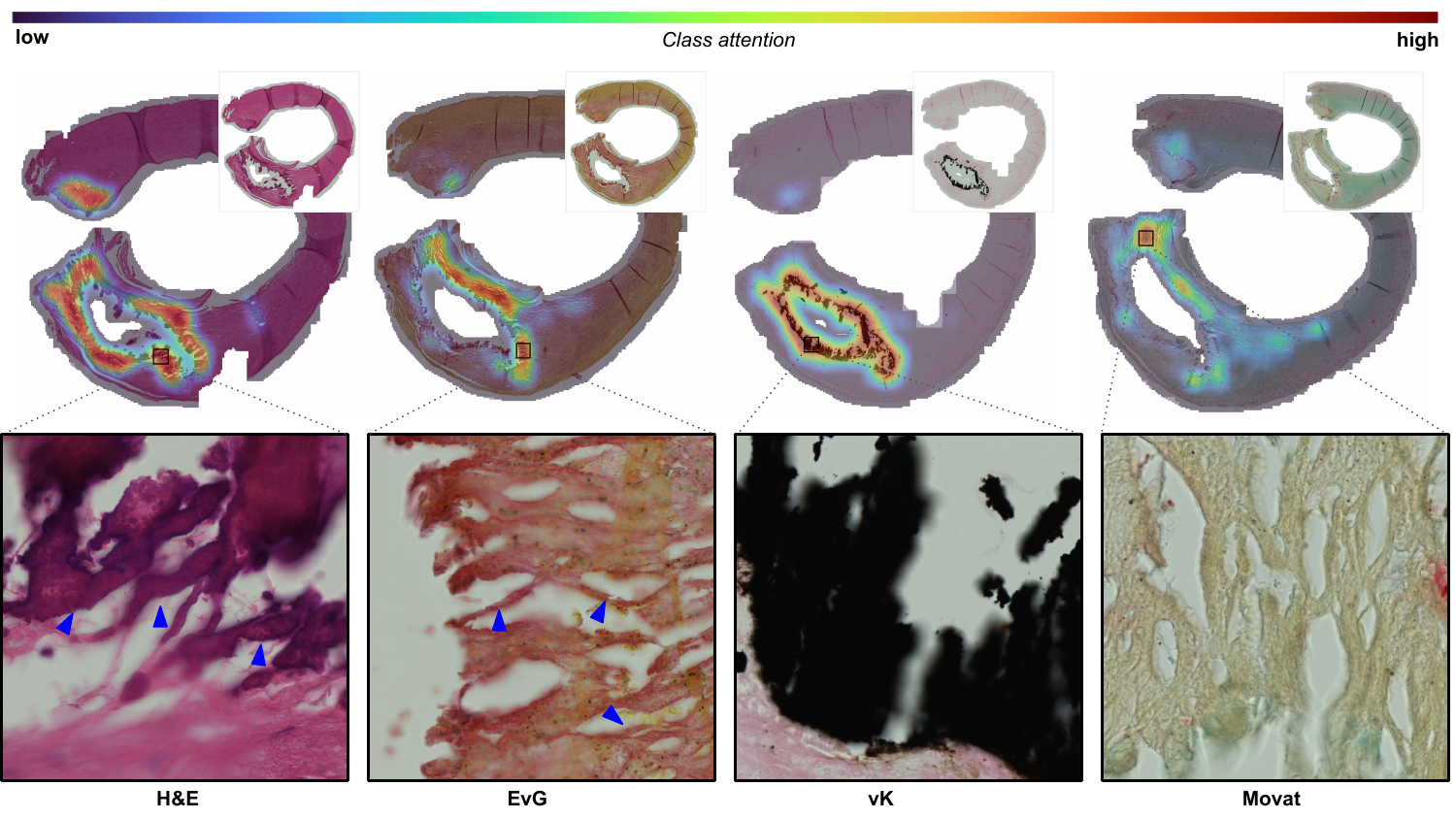}
    \caption{\textbf{UNICORN highlights stain specific classification relevant regions.}\\
Highest class attention regions are shown in red across the four different stainings (H\&E, EvG, vK, Movat) for a calcified fibroatheroma (CFA) case. The  enlarged regions show highest class attention regions. For H\&E and EvG the most relevant structures are highlighted with blue arrows.
}
    \label{fig:fig5}
\end{figure}

Figure \ref{fig:fig5} shows, as an example for the CFA class, class attention maps and corresponding high class attention patches according to figure 3d, revealing regions identified by UNICORN as critical for classification. Additional examples for AIT, PIT, EFA and LFA cases are shown in the appendix (supplementary figure \ref{fig:sfig1}). The high class attention patches in the CFA case demonstrate the most relevant regions for the phenotype learned by the model and closely match the expert opinion. The sections show a lipid-rich necrotic core consisting of extracellular lipid deposits and cellular debris across the four different stains. The necrotic core is covered by a fibrous cap composed of smooth muscle cells, collagen and other extracellular matrix components. 

In the H\&E-stained sections, the algorithm focuses on regions characterized by dense fibrotic areas and calcified necrotic cores (blue arrows), which are typical of advanced fibroatheromas. These areas, highlighted in purple, correlate with expert annotations marking regions of calcification and fibrosis. In the EvG stain, the algorithm detects regions rich in extracellular matrix components, particularly elastic and collagen fibers, shown as red wavy structures (blue arrows). These structures, particularly in the shoulder region, outer necrotic core and intima are critical in distinguishing CFA from earlier lesion stages, as CFA indicates less extracellular matrix due to remodeling in the vessel wall. Von Kossa (vK) staining accentuates calcifications, with the model placing particular attention to black-stained regions, indicating areas of significant calcium deposition within the intima. This focus assists in the differentiation of CFA, where calcification is a defining feature. Finally, the Movat pentachrome stain highlights the lipid-rich necrotic core in yellow-green, which the algorithm correctly identifies as a key pathological feature of CFA. By focusing on these areas, the algorithm demonstrates its ability to distinguish CFA from less advanced fibroatheromas. These visualization results, together with the results shown in figure \ref{fig:fig2}, underscore the algorithm's ability to accurately identify relevant histologic features across multiple staining techniques.

\subsection*{Discussion}
In our study, we present UNICORN, a transformer-based model designed to analyze multi-stain histopathological data. UNICORN represents a significant advancement in histopathology, particularly in integrating multi-modal data for improved diagnostic accuracy. As precision medicine and personalized healthcare advance, the model’s capability to handle missing data and synthesize information from various sources makes it a valuable tool in research and clinical practice. UNICORN might augment pathologist workflows by providing explainable preliminary assessments and highlighting critical areas of interest in tissue samples, thereby expediting the diagnostic process. Its ability to perform inference on only a subset of possible stains can reduce the need for additional staining, optimizing resource use and cost.
The UNICORN framework holds promise for application beyond coronary artery disease. Its architecture is adaptable to other disease phenotypes and could be augmented by integrating high-throughput functional or longitudinal data. This integration would offer deeper insights into disease mechanisms and support more precise phenotyping. Future research could also investigate the replacement of initial expert modules with alternative networks, such as Convolutional Neural Networks or Vision Transformers, to further enhance model performance. Additionally, the incorporation of clinical and high-throughput data into the model could provide a more comprehensive understanding of disease specific pathology, functional biology and improve classification accuracy.
\\
Despite its promising capabilities UNICORN incorporates several limitations. The performance of the model is dependent on the quality and diversity of used staining protocols. In cases where specific stains, such as Movat or Elastica van Gieson, are less informative when other stains are present, the model's effectiveness may be reduced. Additionally, the current feature extractor, optimized primarily for H\&E staining, may not fully leverage information from other stains, potentially limiting the model’s overall performance. Future work should explore fine-tuning strategies and the integration of alternative networks to address this issue. Furthermore, while UNICORN handles missing data well, its generalizability to other phenotypes and the integration of multi-omics data remains to be validated. Finally, the practical implementation in research and clinical settings will require significant investment in training and adaptation of existing workflows. In conclusion, the UNICORN model represents a significant advancement in the integration and analysis of multi-stain histopathological data. Its ability to handle missing modalities and provide comprehensive insights across diverse staining techniques positions it as a valuable tool for enhancing diagnostic accuracy and efficiency. While the model shows promising results in atherosclerosis classification, further optimization and validation are needed to fully realize its potential across various disease phenotypes. The integration of UNICORN into research and clinical workflows has the potential to significantly improve diagnostic consistency and support explainable personalized medicine.
\subsection*{Methods}
\subsubsection*{Histological data compilation, selection and staining}
The Munich cardIovaScular StudIes biObaNk (MISSION) was launched in 2019 and comprises seven cardiovascular relevant tissue samples from each individual, such as coronary and carotid artery tissue samples, as well as myocardium. It also includes blood and plasma, liver, skeletal muscle and various adipose tissues from more than 1000 deceased individuals, collected in formalin for FFPE sections (formalin fixed paraffin embedded)  and fresh frozen at -80°C. tissue sections were deparaffinized in xylene substitute and rehydrated through graded alcohol series. A total of  1045 tissue samples from 170 individuals were analyzed in this study, covering the following coronary arteries: each proximal and distal RCA (right coronary artery), the LAD (left anterior descending artery) and the LCX (left circumflex coronary artery)  as well as the LM (left main) (figure \ref{fig:fig1}a).  Each tissue section was subjected to a series of histopathological analyses, including hematoxylin and eosin (H\&E) staining, Elastica van Gieson (EvG) staining, von Kossa (vK) silver staining, and Movat Pentachrome (Movat) staining. These techniques were employed to assess the histopathological characteristics of atherosclerotic lesions.

The histological classification of the atherosclerotic lesions was performed in accordance with the American Heart Association (AHA) classification and the adapted classification system proposed by Virmani et al. \cite{Virmani2000-jr}. The lesions were categorized into the following five stages (figure \ref{fig:fig1}b). Advanced intima thickening (AIT) is defined by diffuse intimal thickening composed of smooth muscle cells and extracellular matrix in the absence of significant lipid accumulation or inflammatory infiltration. The pathological intima thickening (PIT) is primarily defined by the accumulation of extracellular lipid and inflammatory cells within the intima, though a necrotic core is absent. For the vulnerable stages, early, late, and calcified fibroatheroma (EFA, LFA, and CFA) are distinguished. Early fibroatheroma (EFA) is distinguished by a thin fibrous cap and a developing lipid core whereas the late fibroatheroma (LFA) is more advanced, exhibiting a larger necrotic core and a thicker, though still potentially unstable, fibrous cap. Additionally, LFA displays greater inflammatory involvement. Calcified fibroatheroma (CFA) represents the late stage of fibroatheroma development, with microscopically and macroscopically calcification present within the necrotic core. The fibrous cap may be thickened, and the lesion exhibits reduced cellularity due to extensive calcification.

The institutional review board and ethics committee of the Technical University of Munich, Germany, approved the protocol of the MISSION Biobank (2018-325-S-KK - 22.08.2018). The study was performed in accordance with the provisions of the Declaration of Helsinki and the International Conference on Harmonisation guidelines for good clinical practice. Human data from the MISSION Biobank can be requested by qualified researchers at the German Heart Center Munich.

\subsubsection*{Image patching and feature extraction}
A publicly available pipeline from Wagner et al. \cite{Wagner2023-qu} was used to extract patches and the corresponding features from the WSIs. As the feature extractor, CTransPath \cite{wang2022transformer} was used for each staining. Blurred regions and uncoloured background are excluded using a Canny edge detection algorithm \cite{canny1986computational} with a threshold of one edge to be detected per patch. A resolution of 5x for atherosclerosis prediction was used, as it empirically has shown to work best.
\subsubsection*{Model training}
UNICORN is trained for 30 epochs using an AdamW \cite{Loshchilov2017-fq} optimizer. The learning rate and the weight decay are set to 2.0e-5. Gradients are accumulated across 16 batches with batch size 1 before an update step is made. To increase robustness and force the model to be able to handle missing data well, it is trained with high domain dropout, where domains are randomly masked and not used as input. During training, for each input, one of the input domains is randomly chosen that will not be masked; all other stainings or additional domains have a chance of 0.7 to be masked. As the loss function, standard cross-entropy is used. 
\subsubsection*{Model attention visualization}
To get high resolution attention maps, slides are patched with 95\% overlap between patches, after which features are extracted. This results in 400 distinct feature bags, each representing the full slide. Each of these bags is independently forwarded through the network. The attention that is generated by attention rollout \cite{abnar2020quantifying} on the expert modules for each of the patches for each bag are all saved. Additionally, for each bag, only a single patch is forwarded and let the model predict a class based on one patch only, which results in the class scores. Using these class scores, one can visualize the Multiclass visualization (figure \ref{fig:fig3}b) by multiplying the color corresponding to a certain class by the class probability for a patch. Multiplying the attention scores with the multiclass score of the predicted class then gives the class attention score. In the end, the scores across overlapping patches are averaged for the detailed maps. All scores are normalized between 0 and 1 in the end.

\subsection*{Acknowledgements}
V.K. was supported by the Helmholtz Association under the joint research school “Munich School for Data Science - MUDS”. This work was also supported by the BMBF-funded de.NBI Cloud within the German Network for Bioinformatics Infrastructure (de.NBI) (031A532B, 031A533A, 031A533B, 031A534A, 031A535A, 031A537A, 031A537B, 031A537C, 031A537D, 031A538A). C.M. has received funding from the European Research Council under the European Union’s Horizon 2020 research and innovation program (grant agreement number 866411 and 101113551) and is supported by the Hightech Agenda Bayern. M.v.S. is supported by an excellence grant of the German Center for Cardiovascular Research (DZHK-81X3600506), the German Heart Foundation (Deutsche Herzstiftung e.V.), a Leducq PlaqOmics Junior Investigator Grant and a Junior Research Group Cardiovascular Diseases Grant of the CORONA Foundation (S199/10085/2021). The acquisition of the automated ZEISS AxioScan 7 slide scanning system was supported by the German Research Foundation (DFG-95/1713-1) and the German Center for Cardiovascular Research (DZHK-81Z0600501). H.S. is co-applicant of the British Heart Foundation (BHF)/German Centre of Cardiovascular Research (DZHK)-collaboration (DZHK-BHF: 81X2600522) and the Leducq Foundation for Cardiovascular Research (PlaqOmics: 18CVD02). Additional support has been received from the German Research Foundation (DFG) as part of the Sonderforschungsbereich SFB 1123 (B02) and the Sonderforschungsbereich SFB TRR 267 (B05). Further, we kindly acknowledge the support of the Bavarian State Ministry of Health and Care who funded this work with DigiMed Bayern (grant No: DMB-1805-0001) within its Masterplan “Bayern Digital II” and of the German Federal Ministry of Economics and Energy in its scheme of ModulMax (grant No: ZF4590201BA8). This study was supported by Deutsches CHIP Register e.V. (www.chip-register.de) and funded by the German Center for Cardiovascular Research (DZHK), Berlin, Germany, (DZHK81X2200145 and DZHK-81X2600520). 
\subsection*{Contributions}
VK led the development of the code, model design, and conducted the evaluations. SB was responsible for data collection and annotation. VK and SB were the primary authors of the manuscript and created the figures. VL and SB started the project. MJ provided guidance to SB on classification tasks, and JS and CM offered advisory support to VK. MS and CM oversaw the project’s initiation. All authors contributed to the critical revision and enhancement of the manuscript.

\bibliographystyle{splncs04}
\bibliography{paperpile}
\newpage
\section*{Supplemental material}
\begin{figure}[htbp]
    \centering
    
    \includegraphics[width=\textwidth]{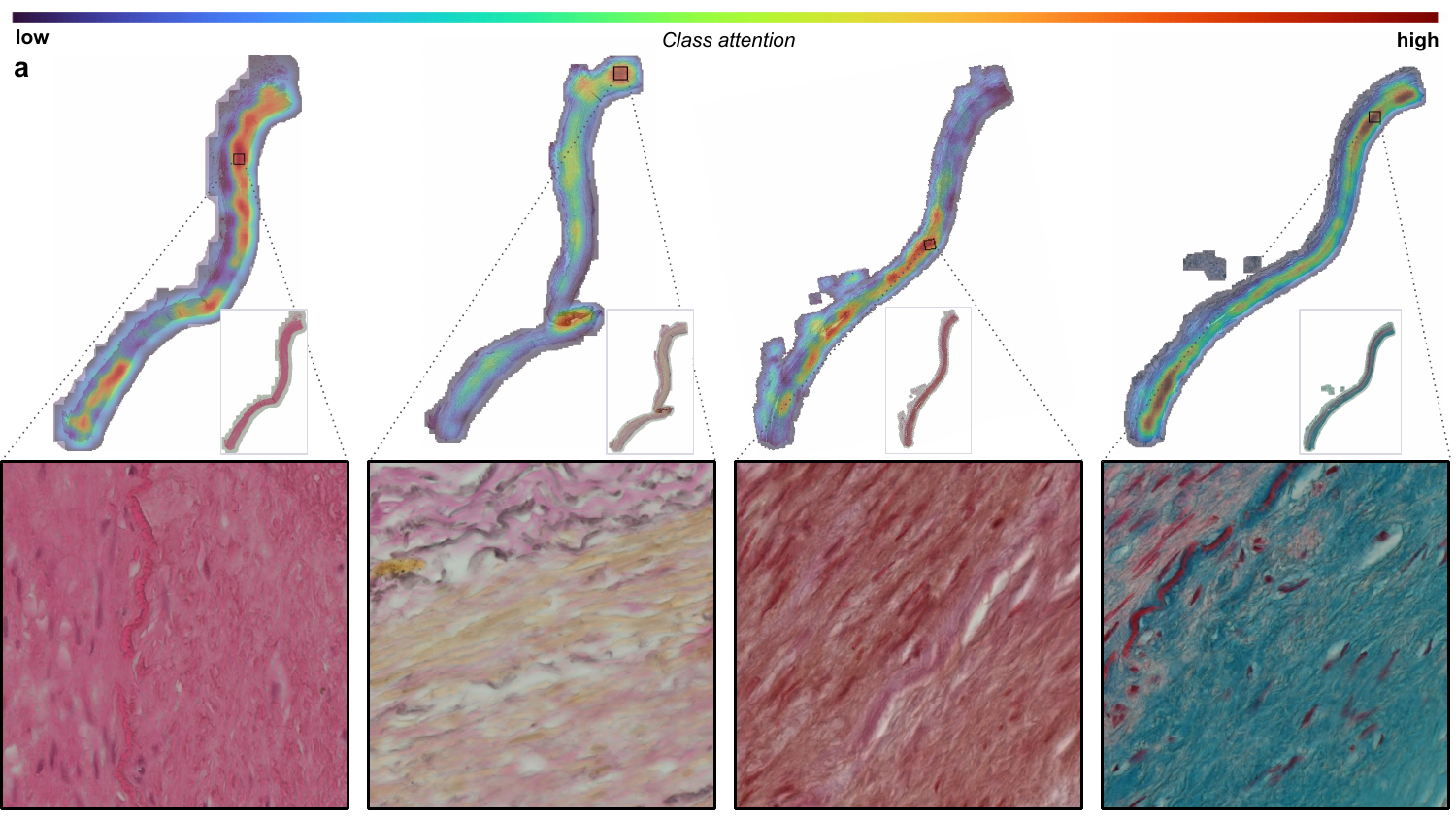}
    \includegraphics[width=\textwidth]{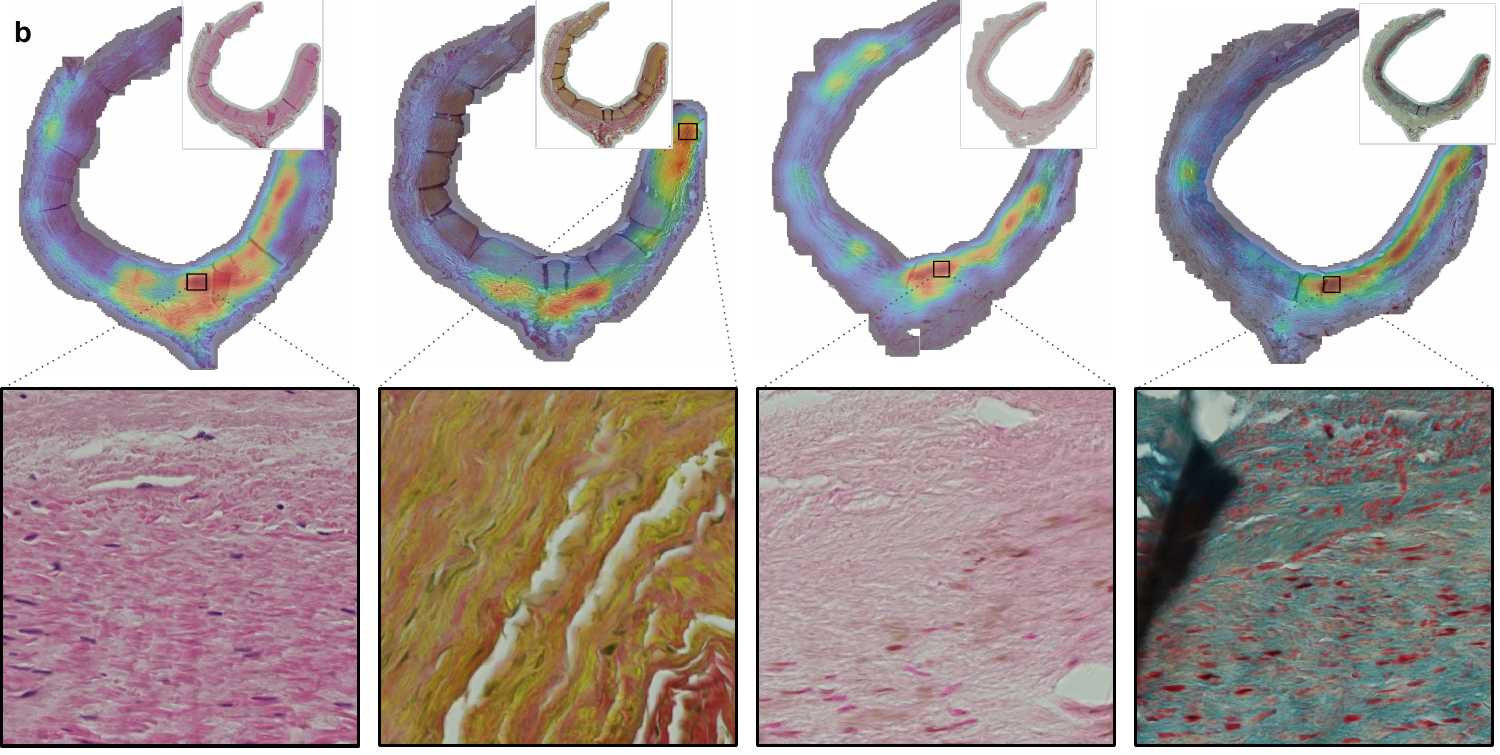}
\end{figure}
\renewcommand{\thefigure}{S\arabic{figure}} 
\setcounter{figure}{0}
\begin{figure}[htbp]
    \centering
    \includegraphics[width=\textwidth]{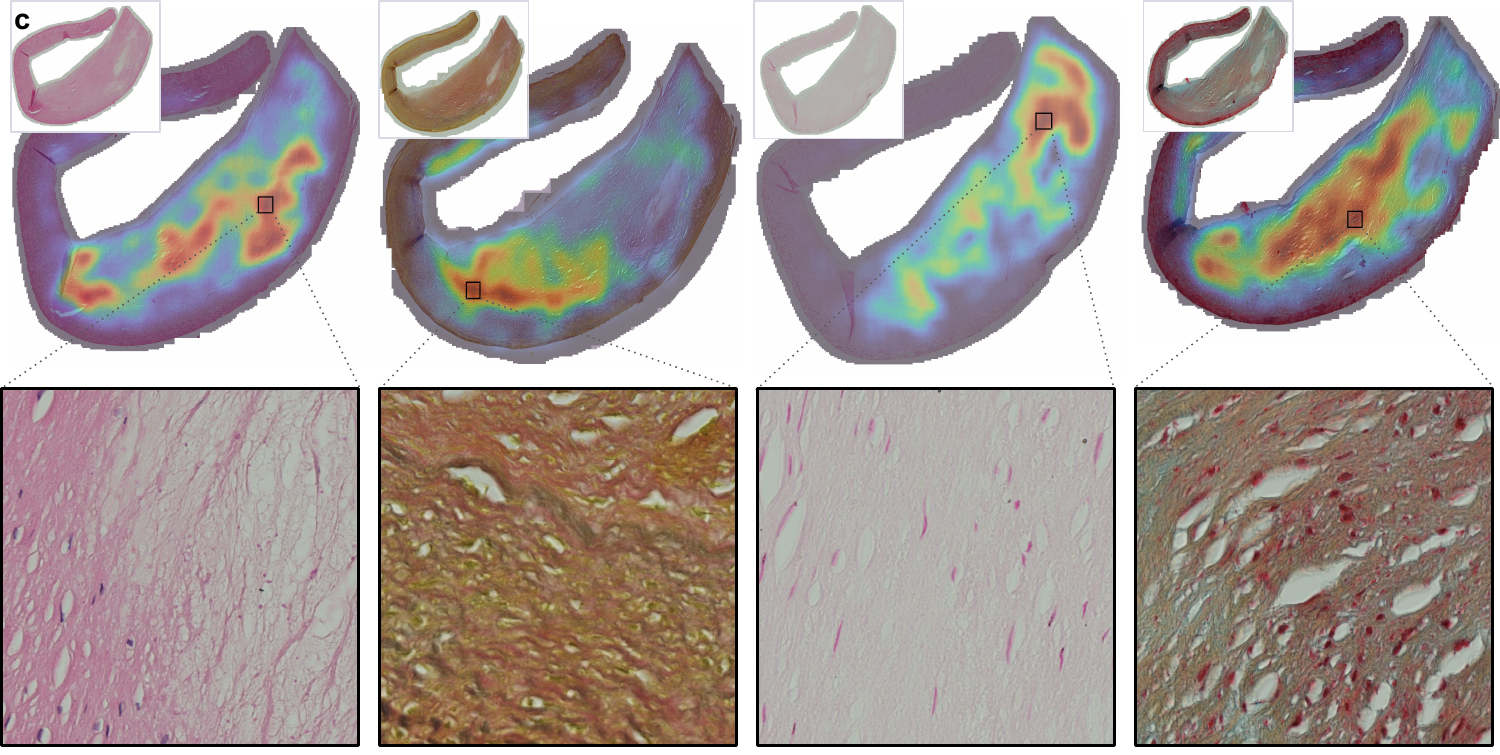}
    \includegraphics[width=\textwidth]{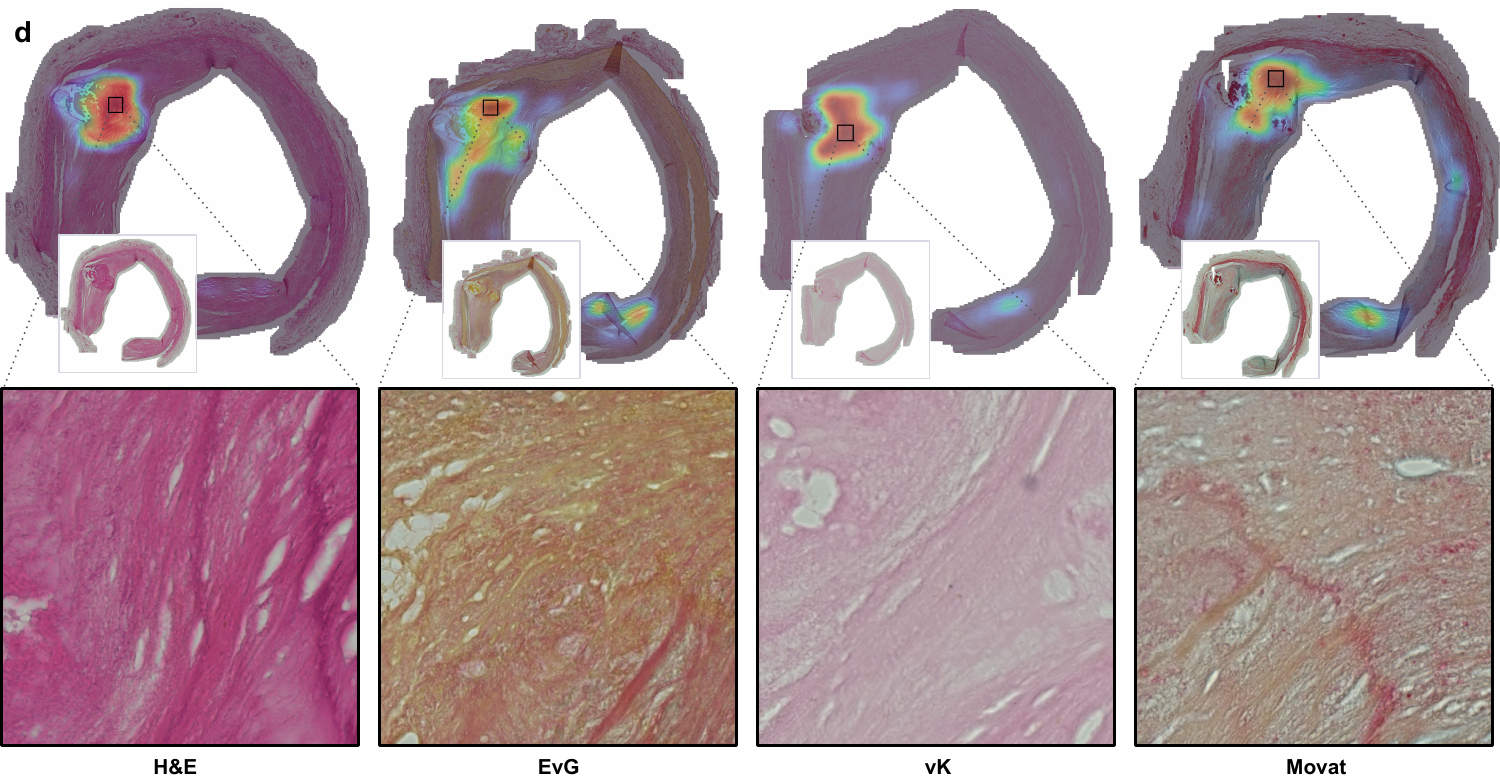}
    \caption{\textbf{UNICORN highlights stain specific classification relevant regions.} \\
    \textbf{a)} UNICORN effectively identifies and classifies adaptive intima thickening (AIT) using four distinct stainings: Hematoxylin and Eosin (H\&E), Elastica van Gieson (EvG), von Kossa, and Movat Pentachrome (Movat). The prioritized patches with the highest score illustrate thickening of the intima with intact cell structures, highlighting the accumulation of smooth muscle cells and extracellular matrix within the arterial wall's intimal layer.\\
    \textbf{b)} Pathologic intima thickening (PIT). The highlighted patches reveal increased numbers of smooth muscle cells within the intima, enhanced deposition of extracellular matrix components such as collagen and proteoglycans, the presence of extracellular lipid pools or droplets within the intimal layer without the formation of a necrotic core, and absent inflammatory cell infiltration. 
    \\
    \textbf{c)} Early fibroatheroma (EFA). The highlighted patches reveal a well-defined lipid-rich necrotic core within the intima, consisting of extracellular lipid deposits and cellular debris. A fibrous cap, composed of smooth muscle cells, collagen, and other extracellular matrix components, covers the necrotic core. There is a presence of macrophages and foam cells at the edges of the necrotic core, along with initial signs of inflammation indicated by some infiltration of inflammatory cells, though not as prominent as in advanced lesions. 
    \\
    \textbf{d)} Late fibroatheroma (LFA). The highlighted patches reveal a prominent, well-defined lipid-rich necrotic core consisting of extracellular lipid deposits, cholesterol crystals, and cellular debris. This core is covered by a thick fibrous cap composed of layers of smooth muscle cells, collagen, and other extracellular matrix components. The presence of macrophages, foam cells, and other inflammatory cells is noted particularly at the edges of the necrotic core and within the fibrous cap. \\
    }
    \label{fig:sfig1}
\end{figure}
\end{document}